\documentclass[letterpaper, 10 pt, conference]{ieeeconf}  %

\IEEEoverridecommandlockouts                              %

\usepackage{graphics} %
\usepackage{epsfig} %
\usepackage{mathptmx} %
\usepackage{times} %
\usepackage{amsmath} %
\usepackage{amssymb}  %
\usepackage[utf8]{inputenc} %
\usepackage[T1]{fontenc}    %
\usepackage{url}            %
\usepackage{booktabs}       %
\usepackage{amsfonts}       %
\usepackage{nicefrac}       %
\usepackage{microtype}      %
\usepackage{xcolor}         %
\usepackage{pifont}
\usepackage{tabularx}
\usepackage{array}
\usepackage{balance}
\usepackage{graphicx}
\usepackage{float}
\usepackage{wrapfig}
\usepackage{caption, subcaption, overpic, textpos}
\usepackage{makecell}
\usepackage{comment}
\usepackage{multicol,multirow}

\usepackage{xspace}
\usepackage[svgnames,table]{xcolor}
\usepackage{algorithm}
\usepackage{algorithmic}
\definecolor{deemph}{gray}{0.6}
\definecolor{baselinecolor}{gray}{.9}
\definecolor{yellow}{RGB}{218,165,32}
\definecolor{lightcyan}{rgb}{0.88, 1.0, 1.0}
\definecolor{lightskyblue}{rgb}{0.53, 0.81, 0.98}
\definecolor{aliceblue}{rgb}{0.94, 0.97, 1.0}
\definecolor{LightSlateBlue}{RGB}{70,130,180}
\definecolor{DeepBlue}{RGB}{65,100,170}
\definecolor{DeepPurple}{RGB}{136,105,160}
\definecolor{LightGreen}{RGB}{59,125,35}
\definecolor{LightRed}{RGB}{234,66,53}
\definecolor{cvprblue}{rgb}{0.21,0.49,0.74}

\usepackage[breaklinks,colorlinks,linkcolor=red,urlcolor=magenta,citecolor=cvprblue,bookmarksopen]{hyperref}

\usepackage[capitalize]{cleveref}
\crefname{section}{Sec.}{Secs.}
\Crefname{section}{Section}{Sections}
\Crefname{figure}{Figure}{Figures}
\crefname{figure}{Fig.}{Figs.}
\Crefname{table}{Table}{Tables}
\crefname{table}{Tab.}{Tabs.}

\makeatletter
\DeclareRobustCommand\onedot{\futurelet\@let@token\@onedot}
\def\@onedot{\ifx\@let@token.\else.\null\fi\xspace}

\def\ie{\emph{i.e}\onedot}

\makeatother

\newcommand{\methodname}{AMS\xspace}

\begin{document}
\title{\LARGE \bf
Agility Meets Stability: \\ Versatile Humanoid Control
with Heterogeneous Data
}

\author{
\authorblockN{
    Yixuan Pan$^{1*}$ \quad 
    Ruoyi Qiao$^{4*}$ \quad 
    Li Chen$^{1}$ \quad
    Kashyap Chitta$^{2}$ 
    \quad Liang Pan$^{1}$ 
    \quad Haoguang Mai$^{1}$ \\
    Qingwen Bu$^{1}$ \quad 
    Hao Zhao$^{3}$ \quad
    Cunyuan Zheng$^{4}$ \quad 
    Ping Luo$^{1}$ \quad 
    Hongyang Li$^{1}$
}
\vspace{0.1em}
\authorblockA{
    $^{1}$The University of Hong Kong  \quad 
    $^{2}$NVIDIA \quad
    $^{3}$Tsinghua University \quad 
    $^{4}$Individual Contributor\\
}
{\small *Equal contribution\quad 
}\\{\texttt{\url{https://opendrivelab.com/AMS/}}}
}

\twocolumn[{%
\renewcommand\twocolumn[1][]{#1}
\maketitle
\vspace{-2.5em}
\begin{center}
    \centering
    \captionsetup{type=figure}
    \includegraphics[width=0.90\textwidth]{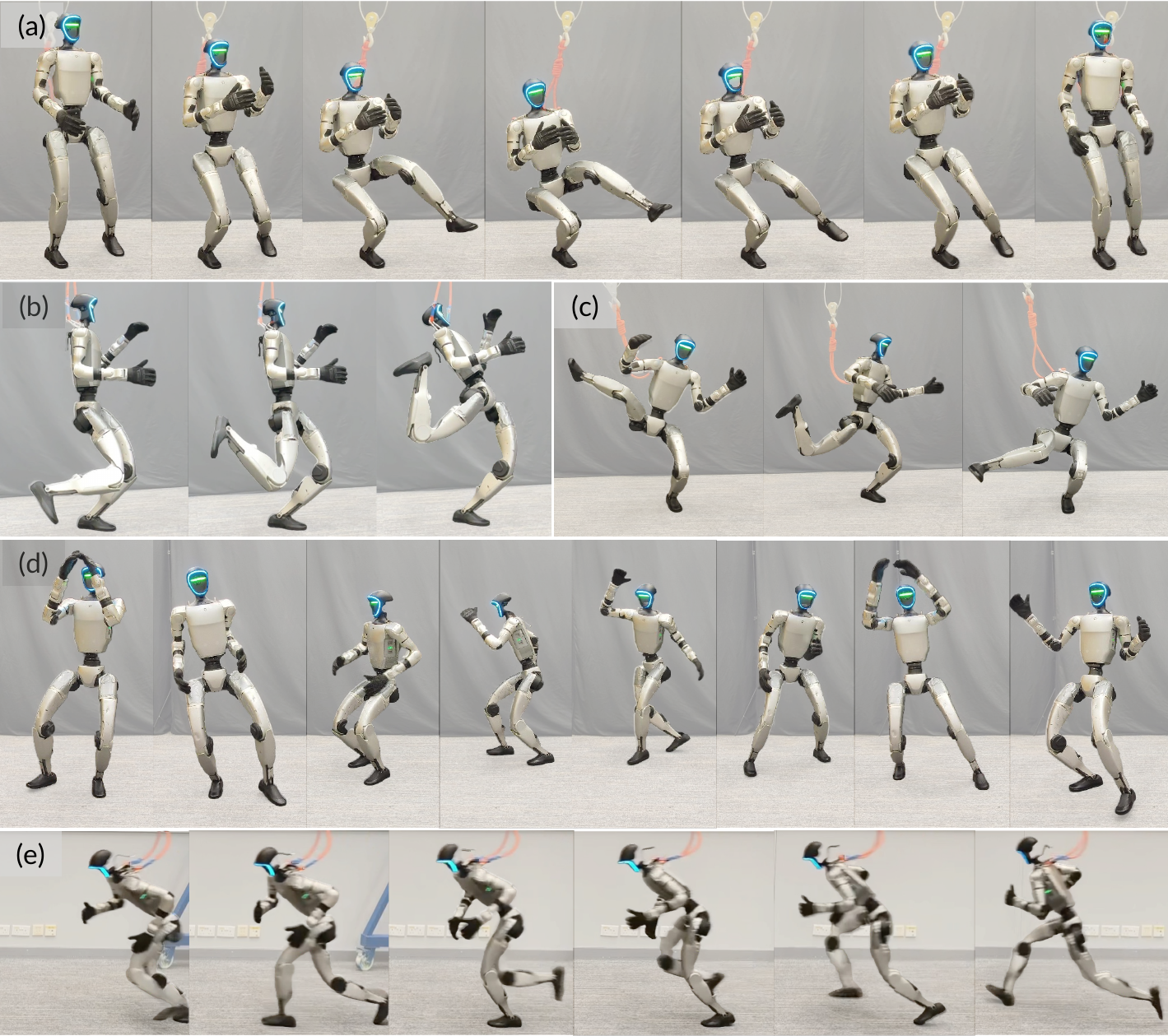}
    \captionof{figure}{ Introducing
    \textbf{\methodname} (Agility Meets Stability), one single policy that 
    performs diverse motions with stability and agility simultaneously on a humanoid robot. 
    The robot can execute challenging balance motions such as (a) \texttt{Ip Man's Squat}, a Kung Fu-style single-leg squat, unseen during training (zero-shot); 
    (b) 
    \texttt{single-leg balance stances} which humans find hard to perform; (c) balanced stretching; 
    as well as expressive motions and high-mobility movements with precise control, such as (d) dancing and (e) running. 
    \label{fig:teaser}}
\end{center}%
}]

\thispagestyle{empty}
\pagestyle{empty}

\begin{abstract}
Humanoid robots are envisioned to perform a wide range of tasks in human-centered environments, requiring controllers that combine agility with robust balance. Recent advances in locomotion and whole-body tracking have enabled impressive progress in either agile dynamic skills or stability-critical behaviors, but existing methods remain specialized, focusing on one capability while compromising the other. In this work, we introduce AMS (Agility Meets Stability), the first framework that unifies both dynamic motion tracking and extreme balance maintenance in a single policy. Our key insight is to leverage heterogeneous data sources: human motion capture datasets that provide rich, agile behaviors, and physically constrained synthetic balance motions that capture stability configurations. To reconcile the divergent optimization goals of agility and stability, we design a hybrid reward scheme that applies general tracking objectives across all data while injecting balance-specific priors only into synthetic motions. Further, an adaptive learning strategy with performance-driven sampling and motion-specific reward shaping enables efficient training across diverse motion distributions. We validate AMS extensively in simulation and on a real Unitree G1 humanoid. Experiments demonstrate that a single policy can execute agile skills such as dancing and running, while also performing zero-shot extreme balance motions like Ip Man's Squat, highlighting AMS as a versatile control paradigm for future humanoid applications.

\end{abstract}

\section{Introduction}
Humanoid robots hold great promise for performing diverse tasks in human-centric environments, from household assistance~\cite{sima2026kai0, shi2025diversity} to industrial applications~\cite{gu2025humanoid, yang2026riseselfimprovingrobotpolicy}. 
Realizing this vision requires robots to emulate the remarkable capabilities that humans naturally master, \ie, versatile, coordinated whole-body skills that seamlessly blend dynamic motion with precise balance. 
Recent progress in locomotion and whole-body tracking has enabled robust outdoor walking~\cite{radosavovic2024humanoid, gu2024advancing}, multi-modal control~\cite{he2025hover, tessler2024maskedmimic}, sequential movements~\cite{zhang2024wococo}, and challenging agile behaviors~\cite{zhuang2024humanoid, ren2025humanoidgoalkeeper, truong2025beyondmimic,huang2025learning}. Despite these advances, humanoid robots still struggle to integrate dynamic motion with precise balance in a \textit{unified} manner.
Humans, by contrast, naturally demonstrate such capabilities, for example, by maintaining a stable single-leg stance while reaching for an object using a free limb as temporary support, or performing precise placement after dynamic walking. 
Endowing humanoids with such integrated versatility, however, remains a fundamental challenge.
Current work typically adopts reinforcement learning (RL) to train whole-body tracking (WBT) policies with human motions as references to accumulate rewards.
They focus on \textit{single-sequence} policy training, either fitting dynamic movements such as ASAP~\cite{he2025asap} or balance motions like HuB~\cite{zhang2025hub}, rather than achieving both capabilities in a unified and generalized manner.

The underlying reasons for this situation can be divided into two aspects: data limitation and divergent optimization objectives.
Existing approaches~\cite{chen2025gmt,he2024omniho,cheng2024exbody} predominantly rely on human motion capture (MoCap) data for training.
While such datasets provide rich dynamic behaviors, they suffer from long-tailed distributions in which extreme balance scenarios are underrepresented.
Further, they inherently restrict the robot to motions that humans can perform, constraining the exploitation of the robot’s unique mechanical capabilities.
In addition, dynamic and balanced motions exhibit distinct distributions and thus require separate optimization objectives.
In an RL-based paradigm, reward functions designed to guide one motion type can inadvertently hinder the learning of the other, leading to conflicts when combined within a unified learning framework. 
For instance, restricting the center of mass to remain above the support foot provides precise guidance for balance tasks but is overly restrictive for dynamic motions that rely on natural momentum transfer and coordinated whole-body movements.
A desirable solution would allow a single policy to learn both dynamic agility and balance robustness without compromising either objective.

To address these challenges, we propose \textbf{\methodname} (Agility Meets Stability), a unified framework that trains a single policy capable of both dynamic motion tracking and extreme balance maintenance through adaptive learning on heterogeneous data. 

Our approach addresses the data limitations by generating constrained synthetic balance motions that complement existing human MoCap data~\cite{mahmood2019amass}. 
Unlike MoCap data, which suffers from sensor noise and kinematic retargeting errors, these synthetic motions are sampled from the humanoid motion space directly while ensuring physical plausibility.
By integrating these heterogeneous data sources, our approach alleviates the long-tailed distribution problem and broadens the range of physically achievable behaviors, complementing and going beyond what traditional human motion datasets can provide.

To resolve conflicting optimization objectives, we employ a hybrid reward scheme that combines two complementary components for policy training. General rewards encourage robust motion tracking across all data, while balance-specific rewards are applied exclusively to the controllable synthetic data, providing precise guidance for stability without inducing conflicts with dynamic tracking objectives.

We further introduce an adaptive learning strategy with two key components for effective learning from heterogeneous data. Adaptive sampling prioritizes challenging motions by automatically adjusting sampling probability for effective hard sample mining. In the meantime, adaptive reward shaping maintains motion-specific error tolerances based on individual performance rather than treating all motions uniformly.

We evaluate \methodname through extensive experiments on a Unitree G1 humanoid robot. As demonstrated in \cref{fig:teaser}, our unified policy effectively executes both dynamic motions such as dancing and challenging balance motions like \texttt{Ip Man's Squat} in a zero-shot manner. 
The versatile framework also enables real-time teleoperation for various motions, highlighting its potential as a foundational control model for autonomous humanoid applications~\cite{jiang2025wholebodyvla, shi2026egohumanoidunlockinginthewildlocomanipulation,chen2025intelligent}.

To summarize, our contributions are threefold: 
\textbf{(1)} We introduce \methodname, the first framework that successfully unifies dynamic motion tracking and extreme balance maintenance in a single policy.
\textbf{(2)} We develop a learning approach that leverages both human-captured motion data and controllable synthetic balance motions, coupled with hybrid rewards and adaptive learning for effective policy training.
\textbf{(3)} We showcase that a single policy can execute both dynamic motions and robust balance control on a humanoid in the real world, outperforming baseline methods and enabling interactive teleoperation.

\begin{figure*}[t]
    \centering
    \includegraphics[width=1.0\textwidth]{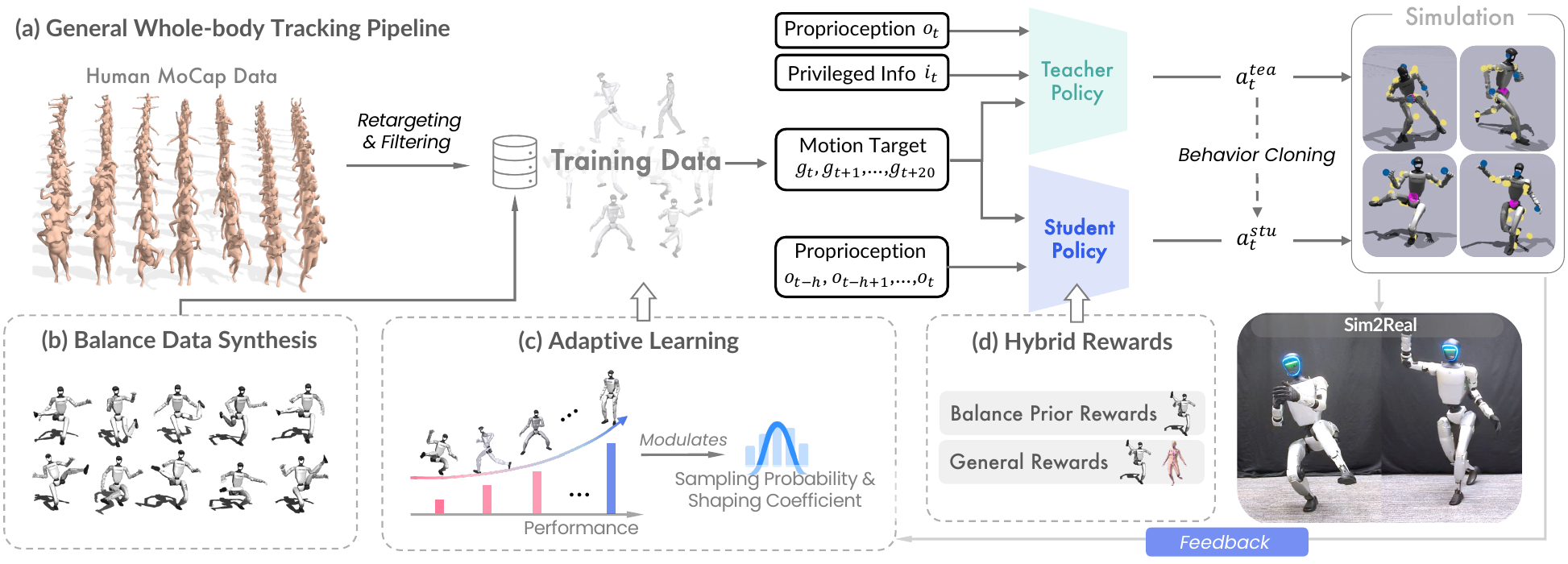}
   \caption{
   \textbf{Overview of \methodname.} 
   \textbf{(a)} The general whole-body tracking pipeline retargets human MoCap data to reference motions and adopts a teacher-student-based strategy for reinforcement learning (\cref{sec:method-wbt}).
   To address data limitations and conflicting optimization objectives, \methodname introduces three key components as follows. \textbf{(b)} Synthetic balance data is generated to complement human MoCap data and address data limitations (\cref{sec:method-synthetic}). 
   \textbf{(c)} Adaptive learning is employed with adaptive sampling and reward shaping based on individual motion performance (\cref{sec:method-adaptive}).
   \textbf{(d)} Hybrid rewards are designed with general rewards for all motions and balance prior rewards exclusively for synthetic motions (\cref{sec:method-hybrid}).
   }
    \label{fig:pipeline}
\end{figure*}

\section{Related Work}

\subsection{Learning-based Humanoid Whole-body Tracking}

Learning-based whole-body tracking has enabled humanoid robots to achieve increasingly versatile behaviors and is promised as a method to collect humanoid data for embodied policy training. Building upon DeepMimic~\cite{peng2018deepmimic}, recent work has demonstrated agile controllers capable of expressive skills such as dancing~\cite{cheng2024exbody,ji2024exbody2}, martial arts~\cite{xie2025kungfubot}, and general athletic maneuvers~\cite{chen2025gmt,ze2025twist}. Other approaches scale to large motion libraries, where universal policies trained on MoCap datasets~\cite{he2024learning, he2024omniho, chen2025gmt} provide broad coverage of human-like movements.

In parallel, another research thrust targets robust balance control, focusing on quasi-static stability rather than agility. HuB~\cite{zhang2025hub}, for example, introduces motion filtering and task-specific rewards to train policies for extreme balancing poses that are typically absent from human datasets. While effective for maintaining stability, such methods often constrain dynamic motions that inherently require momentum and transient instability.

These two directions, agility and stability, have so far been pursued largely in isolation. \methodname aims to bridge this gap by training a single policy on heterogeneous motion data that integrates both dynamic and balance-critical examples, thereby achieving high-fidelity tracking and robust stability within a unified framework.

\subsection{Motion Targets for Policy Learning}

The capabilities of a learned controller are fundamentally shaped by the reference motions it is trained to imitate. 
As with other learning-based tasks~\cite{wu2025freetacman,zhang2026sparse}, whole-body tracking largely relies on human-collected data. Such data typically comes from large-scale motion capture datasets~\cite{mahmood2019amass,ionescu2013human3} or from monocular video through pose estimation~\cite{shin2023wham,shen2024gvhmr}, which provide diverse and natural kinematics for general-purpose policies~\cite{chen2025gmt,ze2025twist,allshire2025visualimitationenablescontextual}. 
However, such data inevitably reflects the bias of human movement and exhibits a long-tail distribution that undersamples balance-critical or robot-specific behaviors~\cite{li2025train}. 
Complementary to this, optimization- and sampling-based approaches generate feasible trajectories directly in the robot’s configuration space, thereby expanding policy coverage toward versatile locomotion~\cite{xue2025unified} and loco-manipulation~\cite{ben2025homie,li2025amo}. 
Recent advances further leverage generative models conditioned on high-level commands, enabling motion synthesis from language or vision-language prompts~\cite{shao2025langwbclanguagedirectedhumanoidwholebody,xue2025leverbhumanoidwholebodycontrol}. Motivated by these prior efforts, our work adopts a heterogeneous data strategy. 
We combine the natural movement of MoCap-derived motions with a controllable generator that produces physically verified, balance-critical behaviors, providing broad supervision for training a single policy that can handle both agility and stability.

\section{Methodology}

\subsection{Problem Setup}
\label{sec:method-wbt}
We formulate humanoid whole-body tracking as a goal-conditioned reinforcement learning (RL) task, where a policy $\pi$ is optimized to track a reference motion sequence in real time. The pipeline is illustrated in \cref{fig:pipeline}(a). Human MoCap data is initially retargeted to the humanoid motion space and erroneous or infeasible motions for humanoids are filtered out~\cite{he2024omniho}. 
At timestep $t$, the system state $s_t$ contains the agent’s proprioceptive observations $o_t$, while $g_t$ denotes the target motion state from reference motions. 
The reward is defined as $r_t = R(s_t, a_t, g_t)$, encouraging alignment between executed and reference motions. The action $a_t \in \mathbb{R}^{23}$ specifies desired joint positions, applied through a PD controller. 
We train a teacher policy with privileged information $i_t$ using Proximal Policy Optimization (PPO)~\cite{schulman2017proximal}, and distill it into a student policy that depends only on deployable sensory inputs with supervision from the teacher policy~\cite{he2024omniho, lee2020learning}.

\begin{figure*}[t]
    \centering
    \includegraphics[width=0.9\textwidth]{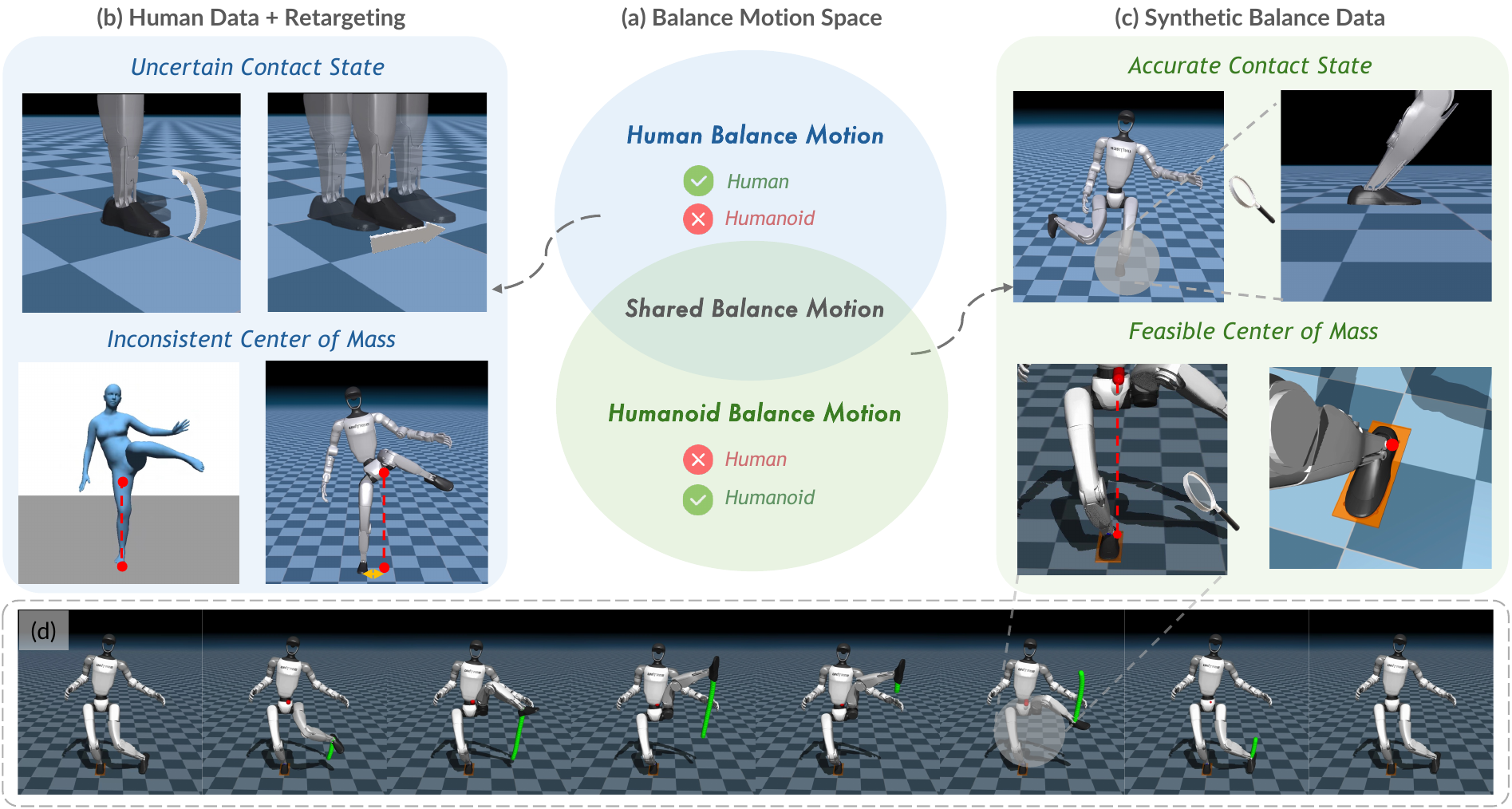}
    \caption{\textbf{Motion space analysis of human data and generated balance data.} 
    \textbf{(a)} Humans and humanoids feature distinctive balance motion spaces, leading to limited reference motions for training whole-body balancing skills. 
    \textbf{(b)} Sensor noise and kinematic retargeting errors greatly affect the reference motion quality from human MoCap data. 
    \textbf{(c)} Constrained synthetic balance data guarantees physical realism, such as the foot contact state and center of mass.
    \textbf{(d)} Example of a generated synthetic balance motion, with the swinging foot trajectory shown in green.}
    \label{fig:generated_data}
\end{figure*}

\subsection{Synthetic Balance Motion Generation}
\label{sec:method-synthetic}

\noindent \textbf{Analysis of Balance Motion References.}
Due to kinematic and morphological differences between humans and humanoid robots, their balance motion spaces only partially overlap, as illustrated in \cref{fig:generated_data}(a). 
Prior works~\cite{he2025asap,chen2025gmt,zhang2025hub,he2024omniho} predominantly rely on human motion data, which inherently constrains the policy's capabilities to this shared space.
However, humanoid robots possess unique mechanical features—different joint limits, actuator capabilities, and mass distributions—that enable balance configurations different from human physical constraints.
Additionally, as shown in \cref{fig:generated_data}(b), human data combined with retargeting introduces noise from sensor measurements and the retargeting process, further limiting the quality of training data.
To address these limitations, we propose generating synthetic balance data by directly sampling from the humanoid balance motion space, as shown in \cref{fig:generated_data}(c) and (d), complementing human-centric datasets with a broader range of feasible behaviors.

\begin{algorithm}[t]
    \caption{Controllable Balance Motion Generation}
    \label{alg:single_motion}
    \renewcommand{\algorithmicrequire}{\textbf{Input:}}
    \renewcommand{\algorithmicensure}{\textbf{Output:}}

    \begin{algorithmic}[1]
        \REQUIRE Robot model $\mathcal{R}$, support foot index $s$, target foot pose $\mathbf{T}_f$, pelvis height $h_p$, horizon $N$, cost weights $\mathbf{\lambda}$
        \ENSURE Motion sequence $\mathcal{M} = \{(\mathbf{X}_t, \mathbf{q}_t)\}_{t=0}^{N-1}$
        
        \textbf{Reference construction:} 
        \STATE \hspace{0.5em} $\mathbf{T}_s(t) \gets \text{constant}(\mathbf{T}_s^0)$ \COMMENT{Support foot trajectory}
        \STATE \hspace{0.5em} $\mathbf{T}_{\bar{s}}(t) \gets \text{interp}(\mathbf{T}_{\bar{s}}^0, \mathbf{T}_f, t/N)$ \COMMENT{Swinging foot trajectory}
        \STATE \hspace{0.5em} $\mathbf{T}_P(t) \gets \text{interp}(\mathbf{T}_P^0, \mathbf{T}_P^{\text{target}}, t/N)$ \COMMENT{Pelvis trajectory}
        
        \textbf{Stage-1 optimization:} 
        \STATE \hspace{0.5em} Solve $\min J_1(\mathbf{X},\mathbf{q})$ where:
        \STATE \hspace{1.0em} $J_1 = \lambda_{\text{track}}\underbrace{\|\mathbf{T}-\mathbf{T}_{\text{ref}}\|_{\mathbf{W}}^2}_{\text{tracking}} + \lambda_{\text{lim}}\underbrace{\text{clip}(\mathbf{q},\mathcal{Q}_{\text{lim}})^2}_{\text{limits}}$
        \STATE \hspace{2.0em} $+ \lambda_{\text{rest}}\underbrace{\|\mathbf{q}-\mathbf{q}^{\text{init}}\|^2}_{\text{rest}} + \lambda_{\text{smooth}}\underbrace{\text{smooth}(\mathbf{X},\mathbf{q})}_{\text{smoothness}}$
        
        \textbf{Stage-2 optimization:} 
        \STATE \hspace{0.5em} Solve $\min J_2(\mathbf{X},\mathbf{q})$ where:
        \STATE \hspace{1.0em} $J_2 = J_1 + \lambda_{\text{bal}}\sum_{t=0}^{N-1} \max(0, d_t - \varepsilon)$
        \STATE \hspace{1.0em} $d_t = \|\max(\mathbf{0}, |\mathbf{p}_t - \mathbf{c}_t| - \mathbf{s})|_2$
        \STATE \hspace{1.0em} $\mathbf{p}_t = \Pi_{xy}\,\text{CoM}(\mathbf{X}_t,\mathbf{q}_t;\mathcal{R})$, $\mathbf{c}_t = \Pi_{xy}\,\text{Trans}(\mathbf{T}_s(t))$
        
        \textbf{Validation:}
        \STATE \hspace{0.5em} \textbf{return} $\mathcal{M}$ if $\max_t d_t \leq \varepsilon$ \textbf{else} \textbf{fail}
    \end{algorithmic}
\end{algorithm}

\smallskip
\noindent \textbf{Motion Generation.} 
To enhance the training dataset with physically plausible and diverse whole-body motion sequences, we propose a motion generation framework that synthesizes balanced whole-body trajectories for single-support maneuvers, as shown in Fig.~\ref{fig:pipeline}(b). The method synthesizes trajectories that transition the ungrounded swinging foot to a target pose while maintaining the center of mass (CoM) within a valid support region, ensuring kinematic feasibility and smoothness.

Given a robot model, a designated support foot, and a time horizon $N$, we first sample a target pose for the swinging foot, a target pelvis height, and an initial joint configuration biased toward natural lower-limb postures with randomized upper-body joints. These samples induce diversity in end-effector goals and whole-body configurations.

We then construct reference trajectories for three key links, including the support foot, swinging foot, and pelvis, using SE(3) interpolation. To compute the motion, we employ a two-stage batch trajectory optimization as outlined in \cref{alg:single_motion}. 

The first stage minimizes a composite cost $J_1$ that includes pose tracking, soft joint limits, rest-pose regularization, and temporal smoothness. This yields a kinematically consistent and smooth trajectory. In the second stage, we augment the cost with a balance-enforcing term: 
\begin{equation}
    J_2 = J_1 + \lambda_{bal} \sum_{t=0}^{N-1}\max(\mathbf{0}, \|\mathbf{p}_t - \mathbf{c}_t - \mathbf{s}\|_2 - \varepsilon),
\end{equation}
where $\mathbf{p}_t$ and $\mathbf{c}_t$ are the 2D projections of the CoM and support foot center, $\mathbf{s} = (s_x, s_y)$ defines the support rectangle, and $\varepsilon$ is a small tolerance. This penalty encourages the CoM to remain within a valid support area. 
The optimization is solved using Levenberg–Marquardt solver~\cite{levenberg1944method, marquardt1963algorithm, nakamura1986inverse,pyroki2025}. Only trajectories that satisfy $\max_t d_t <= \varepsilon$ are accepted, ensuring physical feasibility.

The two-stage approach hierarchically separates kinematic feasibility from balance constraints, where the first stage establishes a robust and smooth trajectory, while the second stage safely refines it for balance, enabling stable convergence.

\subsection{Hybrid Rewards}
\label{sec:method-hybrid}

A central challenge in training a single policy for both dynamic motion tracking and balance-critical behaviors is the conflicting objectives: rewards emphasizing balance could restrict dynamic motions, while rewards for agility may compromise stability. To address this, we introduce a hybrid reward scheme that distinguishes between general motion tracking and balance-specific guidance based on the motion source, as shown in Fig.~\ref{fig:pipeline}(d).

For human motion capture data~\cite{mahmood2019amass,harvey2020robust}, we rely on general motion-tracking reward terms, which encourage natural, human-like movements. In contrast, for synthetic balance-critical motions, we augment the supervision with balance-specific priors, including center-of-mass alignment and foot contact consistency~\cite{zhang2025hub}. These priors provide physically grounded guidance, ensuring a feasible balance without overly constraining the agility.

By selectively applying balance priors rewards only to synthetic data, the hybrid reward design enables the policy to capture agile behaviors from human motions while maintaining reliable stability in challenging postures.

\subsection{Adaptive Learning}
\label{sec:method-adaptive}

To further address both data limitations and conflicting objectives, we introduce an adaptive learning strategy comprising two key components, \ie, adaptive sampling and adaptive reward shaping, as shown in Fig.~\ref{fig:pipeline}(c).

\smallskip
\noindent \textbf{Adaptive Sampling.} We propose a performance-driven adaptive sampling strategy that dynamically adjusts motion sequence sampling probabilities based on tracking performance assessment. Performance is evaluated along three dimensions: (1) execution failure, (2) mean per-joint position error (MPJPE), and (3) maximum joint position error. For each motion $i$ we maintain a probability $p_i$, updated periodically.

Let $\mathcal{F}$ denote the set of failed motions during evaluation, and $e_{mean}^{i}$, $e_{max}^{i}$ represent the mean and maximum joint position errors for motion $i$, respectively. For successful motions, we define performance thresholds using percentiles of the error distribution: $\tau_{poor} = P_{75}(e)$ and $\tau_{good} = P_{25}(e)$, where $P_k$ denotes the $k$-th percentile.

The probability update mechanism operates as follows:
\begin{equation}
p_i^{t+1} = \begin{cases}
p_i^{t} \cdot \gamma_{fail}, & \text{if } i \in \mathcal{F}, \\
p_i^{t} \cdot g_i, & \text{otherwise},
\end{cases}
\end{equation}
where $\gamma_{fail}$ is the failure boost factor, and the adjustment factor $g_i$ is computed as:
\begin{equation}
g_i = 1 + w_{mean}(f_{mean}(e_{mean}^{i}) - 1) + w_{max}(f_{max}(e_{max}^{i}) - 1),
\end{equation}
where $w_{mean}$ and $w_{max}$ are weighting coefficients that control the relative importance of mean and maximum error adjustments. The error-specific adjustment functions $f_{mean}(\cdot)$ and $f_{max}(\cdot)$ are defined identically as:
\begin{equation}
f(e) = \begin{cases}
\beta_{min} + (\beta_{max} - \beta_{min}) \cdot r_{poor}, & \text{if } e > \tau_{poor}, \\
\alpha_{min} + (\alpha_{max} - \alpha_{min}) \cdot (1 - r_{good}), & \text{if } e < \tau_{good}, \\
1, & \text{otherwise},
\end{cases}
\end{equation}
where $\beta_{min}, \beta_{max} > 1$ are the minimum and maximum boost factors for poor-performing motions, $\alpha_{min}, \alpha_{max} < 1$ are the minimum and maximum reduction factors for well-performing motions. The normalized ratios are computed as:
\begin{equation}
r_{poor} = \frac{e - \tau_{poor}}{e_{max} - \tau_{poor}}, \quad r_{good} = \frac{\tau_{good} - e}{\tau_{good} - e_{min}},
\end{equation}
with $e_{max}$ and $e_{min}$ being the maximum and minimum errors observed in the current evaluation.

To ensure exploration and prevent any motion from being completely ignored, we enforce a minimum sampling probability constraint. After updating all probabilities, they are first normalized, then clamped to a minimum threshold:
\begin{equation}
p_i^{final} = \max\left(\frac{p_i^{t+1}}{\sum_{j=1}^{N} p_j^{t+1}}, p_{min}\right),
\end{equation}
where $p_{min} = \lambda \cdot \frac{1}{N}$ with $\lambda$ being the minimum probability factor and $N$ the total number of motions. The probabilities are then re-normalized to ensure they sum to unity.

This adaptive sampling mechanism enables \methodname to automatically focus on poorly-tracked motion patterns by continuously adjusting the training data distribution based on tracking performance, thereby improving both sample efficiency and generalization performance.

\smallskip
\noindent \textbf{Adaptive Reward Shaping.} Existing universal WBT methods~\cite{ze2025twist, he2024omniho, li2025clone, yin2025unitracker} typically employ uniform and fixed shaping coefficients to modulate reward functions for all motions. Typically, the reward is defined as $r = \exp(-\textit{err}/\sigma)$, where $\textit{err}$ represents the tracking error for a given motion and $\sigma$ serves as the shaping coefficient controlling error tolerance. 
However, this uniform treatment presents two challenges: (1) fixed tolerance does not adapt to improving tracking performance; (2) identical parameters create conflicting objectives between dynamic and balance motions that require different shaping strategies.

Inspired by PBHC~\cite{xie2025kungfubot}, we extend their adaptive strategy from single-motion tracking to general multi-motion tracking scenarios. Specifically, we maintain motion-specific $\sigma$ parameter sets, with separate adjustments for different body parts. 
For stable and responsive adaptation, we employ Exponential Moving Average (EMA) to update these parameters:
\begin{equation}
\sigma_{\text{new}} = (1-\alpha) \cdot \sigma_{\text{current}} + \alpha \cdot \text{err}_{\text{current}},
\end{equation}
where $\alpha$ is the 
update rate controlling adaptation responsiveness, and $\text{err}_{\text{current}}$ represents the current tracking error.

This motion-specific adaptive reward shaping mechanism enables \methodname to simultaneously adapt to training progress and motion diversity, significantly improving learning efficiency in general motion tracking scenarios.

\begin{table*}[t]
\centering
\setlength{\tabcolsep}{4pt}
\definecolor{lightblue}{RGB}{173,216,230}
\definecolor{lightgray}{RGB}{240,240,240}
\caption{\textbf{Simulation performance comparison on different datasets and ablation study.} Our method consistently achieves lower tracking errors and higher success rates across both agile motion and challenging balance motions, demonstrating strong generalization and robustness.
}
\label{tab:performance}
\resizebox{\textwidth}{!}{%
\begin{tabular}{lccccccccccc}
\toprule
                                               & \multicolumn{3}{c}{\textbf{MoCap Data (AMASS+LAFAN1)}}                                                                              & \multicolumn{5}{c}{\textbf{Synthetic Balance Data}}                                                                                          & \multicolumn{3}{c}{\textbf{All}}                                                                                      \\ 
\cmidrule(lr){2-4} \cmidrule(lr){5-9} \cmidrule(lr){10-12} 

\textbf{Method}                                         & \multicolumn{2}{c}{Tracking Error}              & \multicolumn{1}{c}{Completion} & \multicolumn{2}{c}{Stability}              & \multicolumn{2}{c}{Tracking Error}              & \multicolumn{1}{c}{Completion} & \multicolumn{2}{c}{Tracking Error}              & \multicolumn{1}{c}{Completion} \\ 
\cmidrule(lr){2-3} \cmidrule(lr){4-4}
\cmidrule(lr){5-6} \cmidrule(lr){7-8} \cmidrule(lr){9-9}
\cmidrule(lr){10-11} \cmidrule(lr){12-12}
                                         & \multicolumn{1}{c}{\cellcolor{lightblue!50}$\mathit{E}_{\text{g-mpjpe}}$ $\downarrow$} & \multicolumn{1}{c}{$\mathit{E}_{\text{mpjpe}}$ $\downarrow$} & \multicolumn{1}{c}{Succ. $\uparrow$} & \multicolumn{1}{c}{\cellcolor{lightblue!50}Cont. $\downarrow$} & \multicolumn{1}{c}{Slip. $\downarrow$} & \multicolumn{1}{c}{$\mathit{E}_{\text{g-mpjpe}}$ $\downarrow$} & \multicolumn{1}{c}{\cellcolor{lightblue!50}$\mathit{E}_{\text{mpjpe}}$ $\downarrow$} & \multicolumn{1}{c}{Succ. $\uparrow$} & \multicolumn{1}{c}{\cellcolor{lightblue!50}$\mathit{E}_{\text{g-mpjpe}}$ $\downarrow$} & \multicolumn{1}{c}{\cellcolor{lightblue!50}$\mathit{E}_{\text{mpjpe}}$ $\downarrow$} & \multicolumn{1}{c}{Succ. $\uparrow$} \\ 

\midrule
\multicolumn{12}{c}{\cellcolor{lightgray}(a) Main Results} \\ 
\midrule
OmniH2O~\cite{he2024omniho}                                        &            \cellcolor{lightblue!50}68.31              &            37.23             &              98.49\%                  &           \cellcolor{lightblue!50}0.24               &             0.038             &            72.51             &           \cellcolor{lightblue!50}56.88              &            99.76\%              &           \cellcolor{lightblue!50}69.84             &      \cellcolor{lightblue!50}44.18                   &                 98.93\%               \\
HuB~\cite{zhang2025hub}                                           &            \cellcolor{lightblue!50}158.80             &           82.13              &             67.03\%                   &            \cellcolor{lightblue!50}\textbf{0.10}              &            \textbf{0.030}              &          137.44               &            \cellcolor{lightblue!50}128.22             &            99.82\%              &         \cellcolor{lightblue!50}151.26                &             \cellcolor{lightblue!50}98.42            &              77.23\%                  \\
\textbf{\methodname (Ours)}                                           &          \cellcolor{lightblue!50}\textbf{48.60}               &          \textbf{24.48}               &             \textbf{99.69\%}                   &            \cellcolor{lightblue!50}0.12              &             \textbf{0.030}             &           \textbf{64.03}               &            \cellcolor{lightblue!50}\textbf{37.30}              &            \textbf{99.95\%}              &            \cellcolor{lightblue!50}\textbf{54.06}             &            \cellcolor{lightblue!50}\textbf{29.02}             &              \textbf{99.78\%}                  \\

\midrule
\multicolumn{12}{c}{\cellcolor{lightgray}(b) Ablation on Synthetic Balance Data} \\ 
\midrule
\methodname w/o Synthetic Balance Data                 &           \cellcolor{lightblue!50}50.25              &                    \textbf{24.10}     &                   99.64\%             &           \cellcolor{lightblue!50}0.69               &           0.047               & 112.20                      &           \cellcolor{lightblue!50}71.89              &           94.54\%               &                 \cellcolor{lightblue!50}72.20               &           \cellcolor{lightblue!50}40.99              &           98.09\%              \\
\textbf{\methodname (Ours)}                                           &          \cellcolor{lightblue!50}\textbf{48.60}               &          24.48               &             \textbf{99.69\%}                   &            \cellcolor{lightblue!50}\textbf{0.12}              &             \textbf{0.030}             &           \textbf{64.03}               &            \cellcolor{lightblue!50}\textbf{37.30}              &            \textbf{99.95\%}              &            \cellcolor{lightblue!50}\textbf{54.06}             &            \cellcolor{lightblue!50}\textbf{29.02}             &              \textbf{99.78\%}                  \\

\midrule
\multicolumn{12}{c}{\cellcolor{lightgray}(c) Ablation on Hybrid Rewards} \\ 
\midrule
\methodname w/ General Rewards Only                     &            \cellcolor{lightblue!50}49.70             &            25.41             &              \textbf{99.72\%}                  &             \cellcolor{lightblue!50}0.39              &            0.036              &             65.39            &           \cellcolor{lightblue!50}45.98              &            99.46\%              &            \cellcolor{lightblue!50}55.31                   &            \cellcolor{lightblue!50}32.75             &         99.65\%                \\
\methodname w/ All Rewards for All Data                     &            \cellcolor{lightblue!50}54.09             &            27.30             &              99.60\%                  &             \cellcolor{lightblue!50}0.31              &            0.095              &             71.62            &           \cellcolor{lightblue!50}40.56              &            99.89\%              &            \cellcolor{lightblue!50}60.32                   &            \cellcolor{lightblue!50}31.99             &         99.70\%                \\
\textbf{\methodname (Ours)}                                           &          \cellcolor{lightblue!50}\textbf{48.60}               &          \textbf{24.48}               &             99.69\%                   &            \cellcolor{lightblue!50}\textbf{0.12}              &             \textbf{0.030}             &           \textbf{64.03}               &            \cellcolor{lightblue!50}\textbf{37.30}              &            \textbf{99.95\%}              &            \cellcolor{lightblue!50}\textbf{54.06}             &            \cellcolor{lightblue!50}\textbf{29.02}             &              \textbf{99.78\%}                  \\

\midrule
\multicolumn{12}{c}{\cellcolor{lightgray}(d) Ablation on Adaptive Learning} \\ 
\midrule
\methodname w/o Adaptive Learning (AS+ARS)               &                \cellcolor{lightblue!50}78.88         &            27.74             &              98.21\%                  &          \cellcolor{lightblue!50}\textbf{0.09}              &           \textbf{0.029}               &           87.86              &          \cellcolor{lightblue!50}43.21               &                \textbf{99.95\%}          &           \cellcolor{lightblue!50}82.11                     &          \cellcolor{lightblue!50}33.22               &            98.75\%             \\
\methodname w/o Adaptive Sampling (AS)                   &            \cellcolor{lightblue!50}52.92             &             24.60            &               98.85\%                 &           \cellcolor{lightblue!50}\textbf{0.09}               &              0.030            &        66.51                 &           \cellcolor{lightblue!50}39.15              &            99.69\%              &             \cellcolor{lightblue!50}57.74                   &         \cellcolor{lightblue!50}29.74                &         99.14\%                \\
\methodname w/o Adaptive Reward Shaping (ARS)               &          \cellcolor{lightblue!50}74.45               &           26.86              &               99.49\%                 &         \cellcolor{lightblue!50}0.13                 &             0.030             &       89.03                  &            \cellcolor{lightblue!50}47.27             &           99.90\%               &              \cellcolor{lightblue!50}79.76                  &           \cellcolor{lightblue!50}34.11              &       99.61\%                  \\
\textbf{\methodname (Ours)}                                           &          \cellcolor{lightblue!50}\textbf{48.60}               &          \textbf{24.48}               &             \textbf{99.69\%}                   &            \cellcolor{lightblue!50}0.12              &             0.030             &           \textbf{64.03}               &            \cellcolor{lightblue!50}\textbf{37.30}              &            \textbf{99.95\%}              &            \cellcolor{lightblue!50}\textbf{54.06}             &            \cellcolor{lightblue!50}\textbf{29.02}             &              \textbf{99.78\%}                  \\

\bottomrule
\end{tabular}}
\end{table*}

\begin{table}[t]
\centering
\setlength{\tabcolsep}{6pt}
\definecolor{lightblue}{RGB}{173,216,230}
\caption{\textbf{Out-of-distribution (OOD) performance comparison.} Our method achieves the lowest tracking errors and highest completion rate, showing better generalization to unseen motions.}
\label{tab:ood}
\resizebox{\linewidth}{!}{%
\begin{tabular}{lccc}
\toprule
\multirow{2}{*}{\textbf{Method}} & \multicolumn{2}{c}{Tracking Error} & \multicolumn{1}{c}{Completion} \\
\cmidrule(lr){2-3} \cmidrule(lr){4-4}
& $\mathit{E}_{\text{g-mpjpe}}$ $\downarrow$ & $\mathit{E}_{\text{mpjpe}}$ $\downarrow$ & Succ.$\uparrow$ \\
\midrule
\methodname w/o Synthetic Balance Data       & 86.61    & 46.43                        & 96.0                              \\
OmniH2O~\cite{he2024omniho} w/ All Data & 76.26    & 49.57                        & 99.1                              \\
\textbf{\methodname (Ours)} & \textbf{63.48} & \textbf{32.06}               & \textbf{99.7}                    \\
\bottomrule
\end{tabular}
}
\end{table}

\section{Experiment}
Our experiments aim to answer the following questions:
\begin{itemize}
    \item \textbf{Q1:} How well does AMS perform on both dynamic and balance motions compared to existing approaches?
    \item \textbf{Q2:} How do the synthetic data and training strategies contribute to the overall performance?
    \item \textbf{Q3:} Can AMS generalize to unseen scenarios and real-world deployment?
\end{itemize}

\subsection{Experimental Setup}  
We evaluate \methodname in both simulation and real-robot experiments. 
In simulation, we use IsaacGym~\cite{makoviychuk2021isaac} as our physics simulator. Our training dataset comprises a filtered subset of the AMASS~\cite{mahmood2019amass} and LAFAN1~\cite{harvey2020robust} datasets, containing over 8,000 motion sequences and 10,000 synthetic balance motion sequences generated by our methods.
For real-world experiments, we deploy our policy on Unitree G1~\cite{unitree2023g1}, a humanoid robot with 23 DoFs and a height of 1.3 meters, weighing about 35kg.

\smallskip
\noindent \textbf{Metrics.}
We evaluate the motion tracking performance using five metrics~\cite{zhang2025hub,he2024omniho}. 
(1) \textbf{Success rate} (Succ., \%). Imitation fails if the average deviation from reference exceeds 0.5m at any point, measuring whether the robot can maintain tracking without losing balance. 
(2) \textbf{Global MPJPE} ($\mathit{E}_{\text{g-mpjpe}}$, $mm$) measures global position tracking accuracy. 
(3) \textbf{Root-relative MPJPE} ($\mathit{E}_{\text{mpjpe}}$, $mm$) evaluates local joint position tracking performance. 
To assess policy stability and fidelity on balance motions, we additionally employ (4) \textbf{Contact mismatch} (Cont., \%), measuring the percentage of frames where foot contact states differ from the reference motion; 
and (5) \textbf{Slippage} (Slip., $m/s$), which quantifies the ground-relative velocity of the support foot, where higher values indicate unstable foot contact.

\subsection{Comparison with Existing Methods}
To address \textbf{Q1}, we compare our method against two representative baselines:

\begin{itemize}
    \item \textbf{OmniH2O~\cite{he2024omniho}} is a general humanoid whole-body motion tracking framework that employs a teacher-student learning paradigm. We adapt OmniH2O to the G1 robot and optimize its curriculum parameters for our experimental setup.
    \item \textbf{HuB~\cite{zhang2025hub}.} Building upon the OmniH2O framework, we re-implement HuB by replacing the reward function with HuB's stability-focused reward design, which emphasizes balance motions and contact-aware tracking.
\end{itemize}

For fair comparison, all baselines are trained from scratch with consistent domain randomization. We evaluate the teacher policies in simulation experiments, and the student policies are derived through direct imitation learning from their respective teachers.
Table~\ref{tab:performance}(a) shows that the proposed method significantly outperforms both OmniH2O and HuB. The approach achieves improvements in tracking performance ($\mathit{E}_{\text{g-mpjpe}}$ and $\mathit{E}_{\text{mpjpe}}$), while simultaneously maintaining high stability (Cont. and Slip.).

\subsection{Ablation Study}
To address \textbf{Q2}, we conduct comprehensive ablation studies on each key component of AMS.

\smallskip
\noindent \textbf{Ablation on Synthetic Balance Data.}

As shown in Table~\ref{tab:performance}(b), the variant \textit{w/o Synthetic Balance Data} demonstrates that training exclusively on MoCap data results in poor performance on balance motions. 
Incorporating synthetic balance data maintains comparable performance on MoCap data while significantly improving tracking performance and stability on balance-critical motions. 
To further answer \textbf{Q3} regarding generalization capability, we collect 1000 unseen motions as out-of-distribution (OOD) test data, including self-recorded random motions and single-leg motions generated by our proposed method. 
As shown in Table~\ref{tab:ood}, adding synthetic balance data to the training set effectively improves OOD performance.

\smallskip
\noindent \textbf{Ablation on Hybrid Rewards.}
As shown in Table~\ref{tab:performance}(c), removing balance prior rewards and using only general rewards (\textit{w/ General Rewards Only}) leads to degraded performance on balance motions, with higher tracking errors and more frequent contact mismatches.
Conversely, applying balance prior rewards uniformly across all motions (\textit{w/ All Rewards for All Data}) shows certain improvement on balance tasks, but creates conflicting objectives that harm overall policy performance.
As evidenced in the table, this approach causes performance degradation on MoCap data, which contains substantial dynamic motions.
Our hybrid reward provides strong balance guidance while avoiding these conflicts, delivering improved overall performance by applying balance-specific rewards exclusively to synthetic data while preserving dynamic motion agility.

\smallskip
\noindent \textbf{Ablation on Adaptive Learning.}
We separately evaluate the two main components of our adaptive learning strategy: Adaptive Sampling (AS) and Adaptive Reward Shaping (ARS), as shown in Table~\ref{tab:performance}(d).
AS adaptively mines hard samples by automatically prioritizing difficult motions, leading to improved success rates and tracking performance on challenging and underrepresented samples.
ARS provides targeted reward adjustments for each individual motion, significantly reducing tracking errors.
When both components are removed (\textit{w/o Adaptive Learning}), the performance degradation is most pronounced, demonstrating that uniform treatment of all motions fails to address the inherent data diversity and difficulty distribution.

\begin{figure}[t]
    \centering
    \includegraphics[width=\linewidth]
    {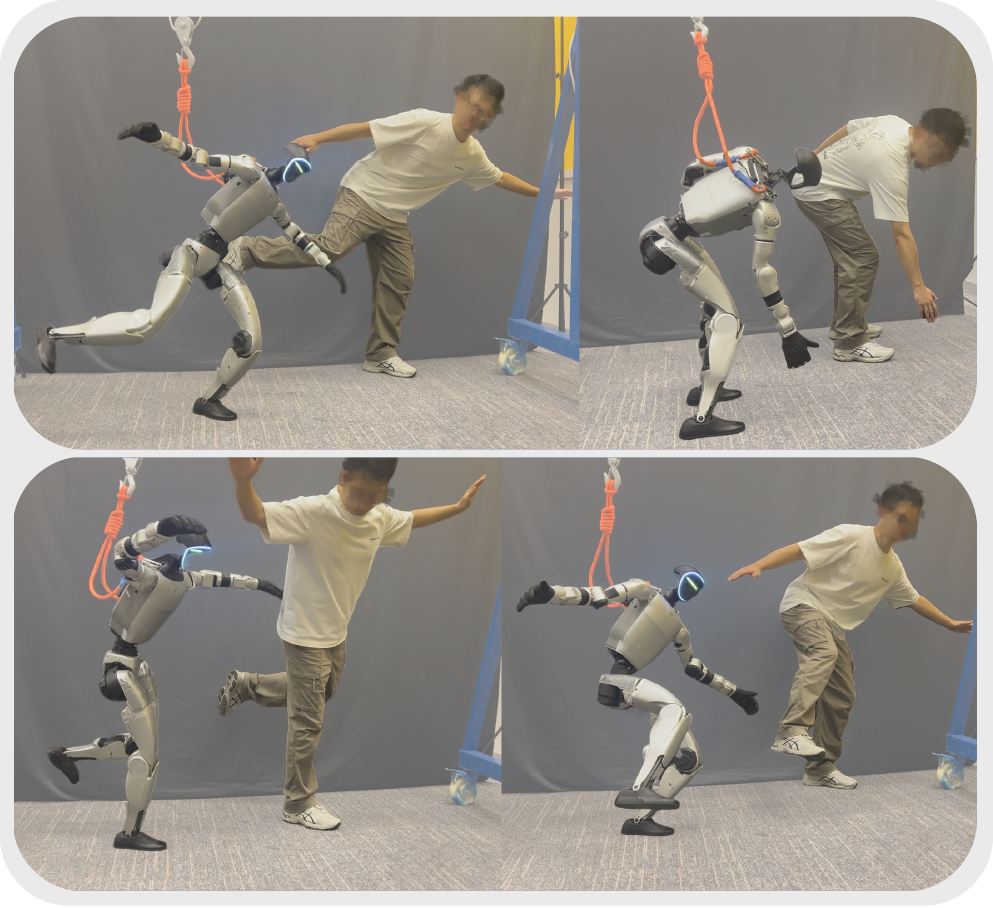}
    \caption{\textbf{RGB camera-based real-time teleoperation.}}
    \label{fig:teleop}
\end{figure}

\subsection{Real-World Deployment}
To further address \textbf{Q3}, we deploy our unified policy on the Unitree G1 humanoid robot, demonstrating execution of a wide range of motions that span both dynamic and balance-critical behaviors.
As illustrated in \cref{fig:teaser}, the robot can execute challenging balancing motions unseen during training, such as \texttt{Ip Man's Squat} and single-leg balancing stances, as well as high-mobility movements and expressive motions like running and dancing.
To further validate generalizability, we conduct real-time teleoperation with an off-the-shelf human pose estimation model~\cite{sarandi2021metrabs}, as shown in \cref{fig:teleop}. The poses' keypoints captured by the RGB camera are scaled to humanoid sizes for tracking. Though not strictly optimized like the complex retargeting process, this simple teleportation still shows robust adaptation to diverse motions.

\section{Conclusion}
In this work, we introduce \methodname, the first framework that successfully unifies dynamic motion tracking and extreme balance maintenance in a single policy. Through leveraging heterogeneous data sources, hybrid rewards, and adaptive learning, our approach enables effective policy training across diverse motion distributions. Our real-world demonstrations showcase that a single policy can execute both dynamic motions and robust balance control, outperforming baseline methods and enabling interactive teleoperation. 

\noindent\textbf{Limitations and Future Work.}
While our approach shows promising results, it lacks precise end-effector control, limiting its applicability to manipulation and contact-rich tasks. Additionally, our RGB-based pose estimation teleoperation system introduces significant noise in global motion estimation, making agile locomotion operations challenging. Future work will explore adopting more precise teleoperation systems, incorporating online retargeting algorithms.

\section*{Acknowledgement}

This study is supported by National Natural Science Foundation of China (62206172). This work is in part supported by the JC STEM Lab of
Autonomous Intelligent Systems funded by The Hong Kong Jockey Club Charities Trust. We are grateful to Jingbo Wang, Shenyuan Gao, and Tairan He for their valuable discussions. We would like to acknowledge Jiacheng Qiu, Shijia Peng, Haoran Jiang, and Zherui Qiu for their assistance and support throughout this project. We also sincerely thank Kinetix AI for supporting the real-world experiments and assisting with demo filming.

{
\bibliographystyle{IEEEtran}
\bibliography{bib_short,ref}
}

\end{document}